\documentclass[journal]{IEEEtran}

\usepackage{graphicx} 
\usepackage{epsfig} 
\usepackage[protrusion=false]{microtype}
\usepackage{times} 
\usepackage{amsmath} 
\usepackage{amssymb}  
\usepackage{bm} 
\usepackage{mathrsfs}
\usepackage{cite}
\usepackage{epstopdf}
\usepackage{subfigure,color}
\usepackage{verbatim}
\usepackage{cases}
\usepackage{booktabs}
\usepackage{mathbbol}
\usepackage{algorithm}
\usepackage{algorithmicx}
\usepackage{algpseudocode}
\usepackage{comment}
\usepackage{gensymb}
\usepackage{textcomp}
\usepackage{threeparttable}
\usepackage{tikz}
\usepackage[colorlinks=true,      
linkcolor=black,      
citecolor=black,      
filecolor=black,      
urlcolor=blue,
bookmarks=false]{hyperref}
\usepackage{array}
\usepackage{diagbox} 
\usepackage{multirow} 
\usepackage{makecell}

\newcommand*{\circled}[1]{\lower.7ex\hbox{\tikz\draw (0pt, 0pt)%
		circle (.5em) node {\makebox[1em][c]{\small #1}};}}

\def\0{{\bf 0}}
\def\1{{\bf 1}}

\title{ 
Orchestrating Wheeled Mobile Robots Online without Conflicts

}

\author{
Jianghong Dong$^{1}$,
Yifeng Zhang$^{2}$, 
Wenhan Cao$^{1}$, 
Jiawei Wang$^{3}$,\\
Mengchi Cai$^{1}$,  
Keqiang Li$^{1}$, and 
Guillaume~Sartoretti$^{2}$ 
	\thanks{Jianghong Dong, Wenhan Cao, Mengchi Cai and Keqiang Li are with the School of Vehicle and Mobility, Tsinghua University, Beijing 100084, China.}
     
\thanks{Yifeng Zhang and Guillaume Sartoretti are with the Department of Mechanical Engineering, National University of Singapore, 117575, Singapore. }
     \thanks{Jiawei Wang is with the Department of Civil and Environmental Engineering, University of Michigan, Ann Arbor, MI 48109, USA. }
}

\begin{document}
\maketitle
\thispagestyle{empty}
\pagestyle{empty}

\begin{abstract}

\looseness=-1
Formation control of wheeled mobile robots (WMRs) has been widely studied due to its broad applications. However, existing studies primarily focus on tracking predefined formations, with limited adaptability to varying environments. To address this limitation, we propose CFOO, a Conflict-Free Online Orchestration framework for continuous formation navigation of WMRs that integrates centralized formation generation with distributed formation maintenance. The upper layer monitors the environment in real time and generates adaptive formations as needed, using the proposed TCF-R2T (Trajectory-Conflict-Free Robot-to-Target assignment) algorithm to compute optimal assignments in polynomial time for timely, conflict-free formation transitions. At the lower layer, each WMR continuously employs JSTP (Joint Spatio-Temporal trajectory Planning) to maintain the given formation by jointly optimizing spatial positions and segment durations, improving coordination over spatial-only optimization with fixed durations. By coordinating these two layers, CFOO orchestrates WMRs to adapt their formations to varying environments and thus enable continuous navigation. Moreover, we design a Time-Varying Formation Reference (TVFR) mechanism to facilitate smooth transitions between successive desired formations. Both simulation and real-world experiments validate the effectiveness and adaptability of CFOO. Experimental videos are available on our project website: \url{https://dongjh20.github.io/CFOO}.

\end{abstract}


\section{Introduction}

\looseness=-1
Formation control of multiple wheeled mobile robots (WMRs) has been extensively studied over the past decades due to its broad applications in areas such as collaborative transportation~\cite{pei2023collaborative}, smart warehouses~\cite{tse2021relative}, resource exploration~\cite{lu2025reinforcement}, and environmental monitoring~\cite{bayat2017environmental}. 
The primary objective of formation control is to maintain desired relative relationships among WMRs. 
This simultaneous coordination in both the lateral and longitudinal directions enables the WMR formation to accomplish complex tasks beyond the capability of a simple collection of individual robots.

\looseness=-1
Existing research on formation control of WMRs has mainly focused on maintaining predefined formations, with relatively few studies addressing formation transitions. 
For formation maintenance, typical approaches include leader--follower~\cite{lin2013leader}, virtual structure~\cite{lewis1997high}, potential-function-based~\cite{de2006formation}, graph-based~\cite{gao2019multi}, and behavior-based methods~\cite{balch1998behavior}. 
These methods generally design feedback controllers based on the tracking error between the current and desired formations~\cite{liu2022formation}, where the error metrics can be broadly categorized into position-based~\cite{lewis1997high}, distance-based~\cite{gao2019multi,de2006formation}, displacement-based~\cite{lin2013leader,wu2024embedded}, and bearing-based~\cite{balch1998behavior}. 
These feedback controllers essentially capture the instantaneous spatial coordination among WMRs.

\looseness=-1
Building on established trajectory tracking controllers, recent studies~\cite{liu2022formation,wu2024embedded} coordinate WMRs at the trajectory level by optimizing spatial positions over fixed segment durations.
Although these methods support convergence to and maintenance of predefined formations, adapting formations to varying environments requires coordinated lateral and longitudinal maneuvers. Jointly adjusting spatial paths and segment durations provides greater flexibility to coordinate when and where WMRs move during these maneuvers. Fixed segment durations restrict this flexibility, potentially delaying formation transitions or compromising feasibility in constrained environments. 
These limitations motivate a framework that enhances spatio-temporal coordination among WMRs and further supports continuous formation navigation through timely formation adaptation.

\begin{figure}[t]
	\centering
    \setlength{\subfigtopskip}{0pt}
    \setlength{\subfigbottomskip}{0pt}
    \setlength{\subfigcapskip}{-3pt}
	\subfigure[Dynamic obstacle avoidance]
	{\includegraphics[width=0.36\textwidth]{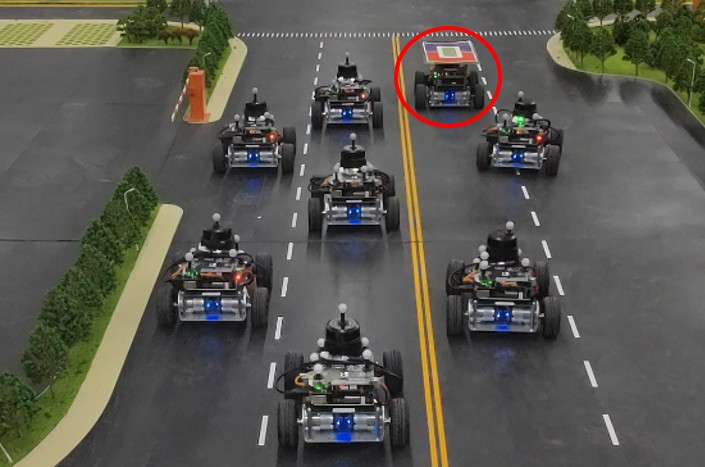}
		\label{fig.exp_obstacle_avoidance} }
    \par\vspace{3pt}
	\subfigure[Formation transition]
	{\includegraphics[width=0.36\textwidth]{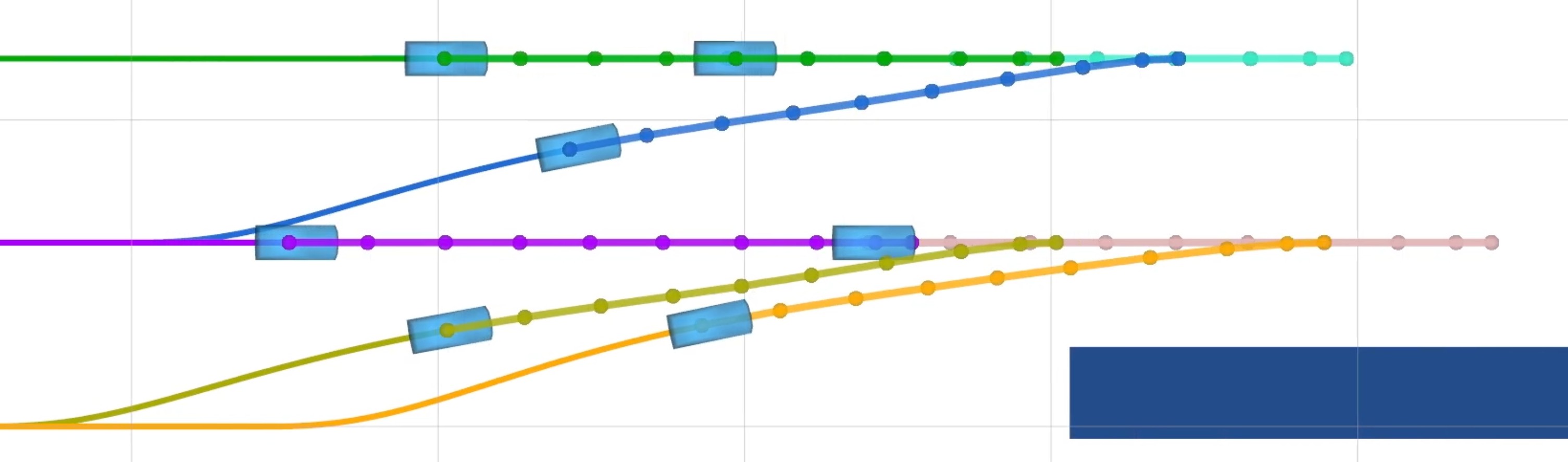}
		\label{fig.exp_rviz_change}}
	\par\vspace{2pt}
	\caption{Handling typical challenges in continuous formation navigation. (a) WMRs avoid a dynamic obstacle (the robot circled in red) while maintaining satisfactory formation performance. (b) The formation promptly transitions from three columns to two as the navigable area narrows.}
    	\label{fig.exp}
	 \vspace{-4mm}
\end{figure}

\looseness=-1
Such adaptation typically requires formation transitions involving both lateral and longitudinal coordination among all WMRs, potentially leading to trajectory conflicts and collisions.
The extent of such conflicts depends strongly on the assignment of WMRs to target positions in the new formation.
Previous studies~\cite{cai2022formation,cai2023formation} typically adopt a decoupled pipeline: a Hungarian-based method is first used to determine the WMR-to-target assignment, followed by A*-based path planning and conflict checking; this process is repeated until all conflicts are resolved and an optimal, conflict-free assignment is obtained.
However, the alternation between assignment attempts and conflict checking reduces computational efficiency, thus motivating the development of a more integrated and efficient solution to support timely formation adaptation when necessary.

\begin{figure}[t]
	\centering
	\includegraphics[width=0.47\textwidth]           {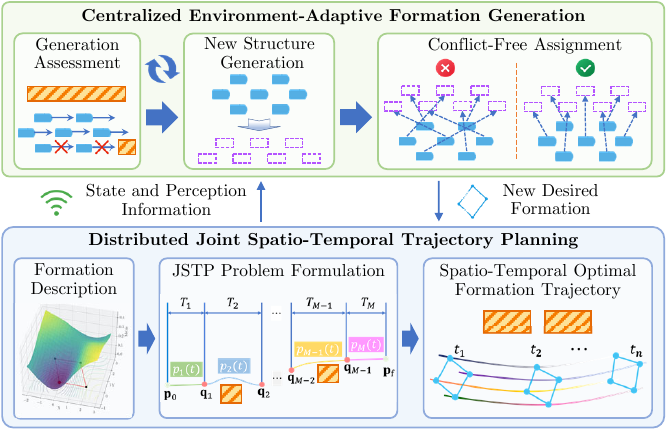}
	\vspace{-2mm}
	\caption{CFOO framework for continuous formation navigation of WMRs. Our upper layer integrates global information and centrally generates environment-adaptive, conflict-free formations when necessary, while each WMR in our lower layer continuously maintains the given formation through distributed trajectory planning.
    }
	\label{fig-REACT-architecture}
  \vspace{-4mm}
\end{figure}

\looseness=-1
In this paper, beyond the extensively studied problem of convergence to and maintenance of predefined formations under various disturbances and constraints, we focus on how to orchestrate WMR formations online to adapt to varying environments, thereby enabling continuous formation navigation. This requires timely generation of environment-adaptive formations and enhanced spatio-temporal coordination to promptly achieve and maintain the desired formation.
In particular, online formation orchestration introduces the challenge of mitigating motion disruptions caused by abrupt changes in the desired formation.
Accordingly, as illustrated in Fig.~\ref{fig-REACT-architecture}, we propose CFOO, a Conflict-Free Online Orchestration framework for continuous formation navigation of WMRs. CFOO integrates TCF-R2T (Trajectory-Conflict-Free Robot-to-Target assignment) for timely, conflict-free formation generation with JSTP (Joint Spatio-Temporal trajectory Planning) for enhanced motion coordination among WMRs.
Moreover, we introduce a Time-Varying Formation Reference (TVFR) mechanism within CFOO to enable smooth transitions between successive desired formations.
Our main contributions are summarized as follows.

\begin{enumerate}
 \setlength\itemsep{0.1em}
	\item Unlike previous studies that mainly focus on tracking predefined formations, we propose CFOO, an online formation orchestration framework for continuous navigation of WMRs in varying environments.
    CFOO leverages centralized formation generation to avoid trajectory conflicts among WMRs during formation transitions, while employing distributed trajectory planning to achieve robust and flexible formation maintenance.
    Moreover, we introduce the TVFR mechanism to enable smooth transitions between successive desired formations.

    \item We propose TCF-R2T, a polynomial-time robot-to-target assignment algorithm for formation generation.
    By constructing a time-expanded network that resolves motion conflicts at each time step, TCF-R2T avoids iterative assignment attempts and conflict resolution required by Hungarian-based methods, improving computational efficiency and enabling timely formation generation.

    \item We propose JSTP, a joint spatio-temporal trajectory planning method for formation maintenance.
    Unlike existing studies that primarily optimize spatial positions under fixed segment durations, JSTP simultaneously optimizes spatial positions and corresponding segment durations, enhancing coordination among WMRs and thus improving formation maintenance performance in the presence of obstacles.
    Moreover, JSTP explicitly incorporates dynamic obstacle avoidance, whereas previous studies primarily consider static obstacles.

\end{enumerate}

\looseness=-1
The rest of this paper is organized as follows. Section~\ref{sec-overview} presents the proposed CFOO framework and the design of its TVFR mechanism. 
Section~\ref{sec-assignment} introduces the TCF-R2T assignment algorithm for conflict-free formation generation, while Section~\ref{sec-optimization} details the JSTP planning method for robust formation maintenance.
Section~\ref{sec-experiment} presents comparisons with existing methods, as well as simulation and real-world experiments. Finally, Section~\ref{sec-conclusion} concludes this paper.

\section{Overview of the CFOO Framework}
\label{sec-overview}

\looseness=-1
This paper aims to develop a flexible and adaptive architecture for continuous navigation of WMRs in varying environments. 
Continuous formation navigation of WMRs essentially involves two aspects: maintaining the current formation and performing formation transitions when required by the environment. 
These transitions should be smooth to avoid abrupt changes in WMR motion.
Accordingly, as shown in Fig.~\ref{fig-REACT-architecture}, CFOO adopts a hierarchical architecture:
(1) upper-layer centralized formation generation by the formation manager, (2) lower-layer distributed trajectory planning by individual WMRs. The formation manager centrally generates environment-adaptive formations when needed, while the WMRs continuously track the generated formation via distributed trajectory planning. 
Under this architecture, tracking a newly generated formation naturally constitutes a formation transition process.
CFOO smooths this transition through the TVFR mechanism.
In the following, we first present the mathematical description of the WMR formation, and then introduce the pipeline of CFOO.

\subsection{Graph-based Formation Description}
\label{subsec-description}

\looseness=-1
Before introducing formation generation and tracking, we first establish the mathematical description of the WMR formation. A formation configuration is composed of two parts: the formation geometry, represented by a set of desired positions, and the robot-to-position assignment, which specifies which WMR is assigned to each position in the formation.
The formation of $N$ WMRs is characterized by a directed graph $\mathcal{G} = (\mathcal{V}, \mathcal{E})$, where $\mathcal{V} := \{1, 2, \ldots, N\}$ and $\mathcal{E} \subseteq \mathcal{V} \times \mathcal{V}$ represent the sets of vertices and edges, respectively. 
Each vertex $u \in \mathcal{V}$ corresponds to a WMR with position vector 
$\mathbf{p}_u = [x_u, y_u]^{\top}$. 
A directed edge $(u,v) \in \mathcal{E}$ from $u$ to $v$ indicates that WMR $v$ can access the relative distance and trajectory information of WMR $u$. 
The weight associated with edge $(u,v)$ is defined by the squared weighted Euclidean distance
$w_{uv} = \| \mathbf{W}(\mathbf{p}_u - \mathbf{p}_v)\|_2^2$, where $\mathbf{W} := \text{diag}(a, 1)$ is the weight matrix and $\|\cdot\|_2$ denotes the 2-norm. The parameter $a$ 
allows certain directions to be prioritized over others when spatial importance varies.

\looseness=-1
Since the primary objective of formation control is to maintain the desired relative relationships among WMRs, 
we naturally adopt the graph representation matrix, i.e., the Laplacian matrix $\tilde{\mathbf{L}}$, which encodes the relative distances among all WMRs, to characterize the formation.
Given the positions $\mathbf{p}_1, \mathbf{p}_2, \ldots, \mathbf{p}_N$, the pairwise edge weights $w_{uv}$ can be computed accordingly. The adjacency matrix is then defined as $\tilde{\mathbf{A}} =[w_{uv}]$,
and the degree matrix is given by $\tilde{\mathbf{D}}=\mathrm{diag}(d_1,\ldots,d_N)$, where $d_u=\sum_{v=1}^{N} w_{uv}$. The Laplacian matrix is  obtained as $\tilde{\mathbf{L}}=\tilde{\mathbf{D}}-\tilde{\mathbf{A}}$.
Therefore, relative positions between WMRs are sufficient for the formation description.


\subsection{Pipeline of CFOO}

\looseness=-1
To ensure consistent formation tracking across all WMRs and avoid conflicts during formation transitions, the manager centrally generates formations and coordinates the WMRs, thereby promoting global efficiency. 
The manager can be deployed either on an external host or on one of the WMRs. 
Meanwhile, each WMR performs distributed trajectory planning independently, which endows the overall framework with flexibility and robustness. 

\looseness=-1
The manager evaluates incoming environmental information in real time to determine whether a new formation is needed.
Changes in the navigable area can render the current formation unsuitable, triggering the generation of a new formation.
A geometric formation structure is first generated, followed by a conflict-free robot-to-target assignment. 
For formation structure generation, we assume that the inter-WMR spacing in the longitudinal ($x$) direction is uniform and adopt the interlaced scheme~\cite{cai2023formation}, which offers improved navigation safety.
Based on the navigable width, target positions are allocated uniformly and symmetrically across available columns, yielding their relative positions directly.
The assignment computation via TCF-R2T is detailed in Section~\ref{sec-assignment}. 
Finally, the generated environment-adaptive formation is dispatched to the WMRs for tracking. 
On the robot side, each WMR executes JSTP to continuously maintain the given formation through joint spatio-temporal optimization, as elaborated in Section~\ref{sec-optimization}.

\looseness=-1
Abrupt updates to the desired formation can cause sudden changes in the formation cost, inducing aggressive corrections and oscillatory motion.
Each WMR employs the TVFR mechanism to mitigate these effects by smoothly interpolating from the current relative configuration to the assigned target configuration:
\begin{equation}
\begin{aligned}
\mathbf{r}_u(t)
&= \mathbf{q}_u^{0}
+ \left(10\xi^{3}-15\xi^{4}+6\xi^{5}\right)
  \left(\mathbf{q}_u^{1}-\mathbf{q}_u^{0}\right),\\
\xi &= \operatorname{clip}\,\!\bigl((t-t_s)/T_s,\,0,\,1\bigr).
\end{aligned}
\label{eq:smooth_formation_reference}
\end{equation}
Here, $\mathbf{q}_u^{0}$ and $\mathbf{q}_u^{1}$ denote the initial and assigned target positions of WMR $u$, each relative to its corresponding formation centroid; $\xi$ is the normalized transition progress, $t_s$ is the locally recorded transition start time, and $T_s$ is the prescribed transition duration. The quintic interpolation yields a $C^2$-continuous reference with zero first and second derivatives at both endpoints. JSTP evaluates its formation cost against the evolving reference, promoting smooth, coordinated transitions during continuous navigation.

\begin{figure}[t]
	\centering
	\includegraphics[width=0.42\textwidth]           {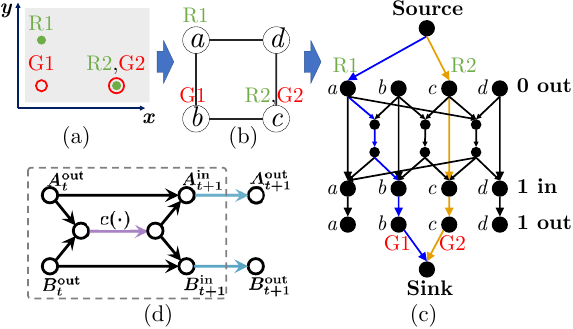}
	\vspace{-3mm}
	\caption{Illustration of the TCF-R2T algorithm. (a) Longitudinally aligned start positions $\mathrm{R1}$, $\mathrm{R2}$ and target positions $\mathrm{G1}$, $\mathrm{G2}$. (b) Undirected graph $G$ constructed from the geometric topology of the start and target positions. (c) Time-expanded network $\mathcal{N}^T$ constructed from graph $G$. (d) The edge-expansion pattern for each edge in the graph $G$, used to connect vertex copies across multiple time steps in the time-expanded network $\mathcal{N}^T$.}
	\label{fig-TCF-R2T}
  \vspace{-2mm}
\end{figure}

\begin{algorithm}[t]
\caption{TCF-R2T}
\label{alg-TCF-R2T}
\begin{algorithmic}[1]
\footnotesize
\Statex \textbf{Input:} Robot positions $\mathbf{R}$, target positions $\mathbf{G}$, cost function $c(\cdot)$ 
\Statex \textbf{Output:} Optimal assignment $\mathcal{A}^*$, conflict-free trajectories $\mathcal{T}^*$

\State $\mathbf{R}', \mathbf{G}' \gets$ longitudinal-alignment\,($\mathbf{R}, \mathbf{G}$)
\State $G=(V,E) \gets$ construct an undirected graph from the geometric topology of $\mathbf{R}'$ and $\mathbf{G}'$ using linear indexing
\State $T_{\min} \gets 0$
\State $T_{\max} \gets N+l$
       \Comment{Exclusive search limit}
\For{$T$ from $T_{\min}$ to $T_{\max}-1$}
    \State $\mathcal{N}^{T} \gets$ construct the $T$-step time-expanded network from $G$ via the edge-expansion pattern with the edge costs given by $c(\cdot)$ 
    \If{a flow $f_T^*$ of value $N$ is obtained on $\mathcal{N}^{T}$}
        \State $\mathcal{A}^*, \mathcal{T}^* \gets$ decode $f_T^*$ via inverse linear indexing
        \State \textbf{break}
    \Else 
        \State \textbf{continue}
    \EndIf
\EndFor
\State \Return $\mathcal{A}^*, \mathcal{T}^*$
\end{algorithmic}
\end{algorithm}

\section{Trajectory-Conflict-Free Robot-to-Target Assignment for Formation Generation}
\label{sec-assignment}

\looseness=-1
In this section, we present TCF-R2T, which assigns WMRs to target positions in the new formation structure and finalizes formation generation, facilitating conflict-free transitions to the assigned target positions.
We first illustrate its workflow  with an intuitive example and then analyze its computational complexity.

\looseness=-1
Given a new formation structure, let $\mathbf{R}$ and $\mathbf{G}$ denote the sets of robot and target positions, respectively. 
For the formation transition from the current formation to a newly generated formation, WMR-to-target assignment aims to minimize the total travel distance while avoiding trajectory conflicts.
We also seek to minimize column changes, reducing lateral maneuvers and potential trajectory conflicts.

\looseness=-1
The workflow of TCF-R2T is summarized in Algorithm~\ref{alg-TCF-R2T}, and an illustrative example is shown in Fig.~\ref{fig-TCF-R2T}.
First, as illustrated in Fig.~\ref{fig-TCF-R2T}(a), since translation preserves the optimal assignment~\cite{agarwal2018simultaneous}, we align the minimum coordinates of the WMRs and targets along the WMRs' forward direction ($x$-direction) to remove redundant translational offsets and thus accelerate computation. 
We then round the coordinates to map the assignment problem onto a grid map. 
Based on the $4$-neighbor geometric topology, we further construct an undirected graph $G=(V,E)$, where vertices $V$ represent the grid points and edges $E$ denote their connections; see Fig.~\ref{fig-TCF-R2T}(b) for illustration. 
The vertex index can be computed directly from its coordinates using linear indexing, i.e., $\mathrm{index} = x + y\bigl(\max(x)+1\bigr)$,  where the coordinate pair can also be uniquely recovered from the index via inverse mapping. 
Resolving trajectory conflicts requires more than assigning WMRs to target positions. We therefore formulate the WMR-to-target assignment problem from a path-planning perspective.

\looseness=-1
To this end, we construct the $T$-step time-expanded network $\mathcal{N}^T$ shown in Fig.~\ref{fig-TCF-R2T}(c).
Specifically, the vertices of graph $G$ are duplicated across $2T+1$ layers, including the initial layer ``$0\,{\mathrm{out}}$'', and for each time step $t$, the beginning layer ``$t\,{\mathrm{in}}$'' and the ending layer ``$t\,{\mathrm{out}}$''. 
For each edge $(A,B)\in E$, the corresponding copies of $A$ and $B$ in adjacent time layers are connected according to the edge-expansion pattern presented in Fig.~\ref{fig-TCF-R2T}(d). 
A ``Source'' node and a ``Sink'' node are then added to connect all start and target vertices, respectively, yielding the structure of $\mathcal{N}^T$.

\looseness=-1
Conflict resolution is achieved by the edge-expansion pattern in Fig.~\ref{fig-TCF-R2T}(d).
Trajectory conflicts arise from motion conflicts at individual time steps, which include edge conflicts and vertex conflicts~\cite{stern2019multi}.
We assign unit capacity to all edges.
The gray dashed box in Fig.~\ref{fig-TCF-R2T}(d) resolves edge conflicts, since unit capacities prevent two WMRs from traversing the same edge in opposite directions at the same time. 
The two cyan edges outside the box resolve vertex conflicts, since unit capacities ensure that at most one WMR can arrive at a vertex at each time step. 
Since conflicts are resolved at every time step, collision-free trajectories are guaranteed throughout the discrete planning horizon. 
For the edge cost $c(\cdot)$, only the purple edge in Fig.~\ref{fig-TCF-R2T}(d) has nonzero cost: $\Delta x_{\max}$ for motions causing a column change, $1$ for motions without column change, and $0$ for waiting, where $\Delta x_{\max}$ is the maximum possible longitudinal travel distance. 
This cost design satisfies our objective of minimizing column changes while also minimizing the total travel distance.
After assigning the capacities and costs, we complete the construction of $\mathcal{N}^T$.

\looseness=-1
We solve this network flow problem on $\mathcal{N}^T$ using the successive shortest path algorithm, obtaining an optimal solution in polynomial time~\cite{ma2016optimal}.
The estimated initial search horizon $T_{\min}$ can be set to $0$. 
The number of algorithm invocations is bounded by $T_{\max}$, whose upper bound is $N+l$, where $N$ is the number of WMRs and $l=\max_{r\in\mathbf{R}',\,g\in\mathbf{G}'}
\operatorname{dist}_{G}(r,g)$ is the maximum shortest-path distance between a start and a target vertex~\cite{yu2013multi}.
Since $l\leq |V|-1$ for a connected graph, where $|V|$ is the number of vertices, $T_{\max}$ is polynomially bounded. 
Therefore, TCF-R2T runs in polynomial time. 
The optimal solution of $\mathcal{N}^T$ is decoded through inverse linear indexing to obtain the optimal conflict-free assignment $\mathcal{A}^*$ and the collision-free grid-based trajectories $\mathcal{T}^*$. 
The resulting trajectories $\mathcal{T}^*$ are then used to initialize the joint spatio-temporal trajectory optimization in Section~\ref{sec-optimization}.

\section{Joint Spatio–Temporal Trajectory Planning for Formation Maintenance}
\label{sec-optimization}

In this section, we present the JSTP method executed by each WMR for formation maintenance.
We first introduce the adopted trajectory representation, then formulate the joint spatio-temporal trajectory optimization problem, and finally present the cost function design and gradient derivation.

\begin{figure}[t]
	\centering
	\includegraphics[width=0.4\textwidth]           {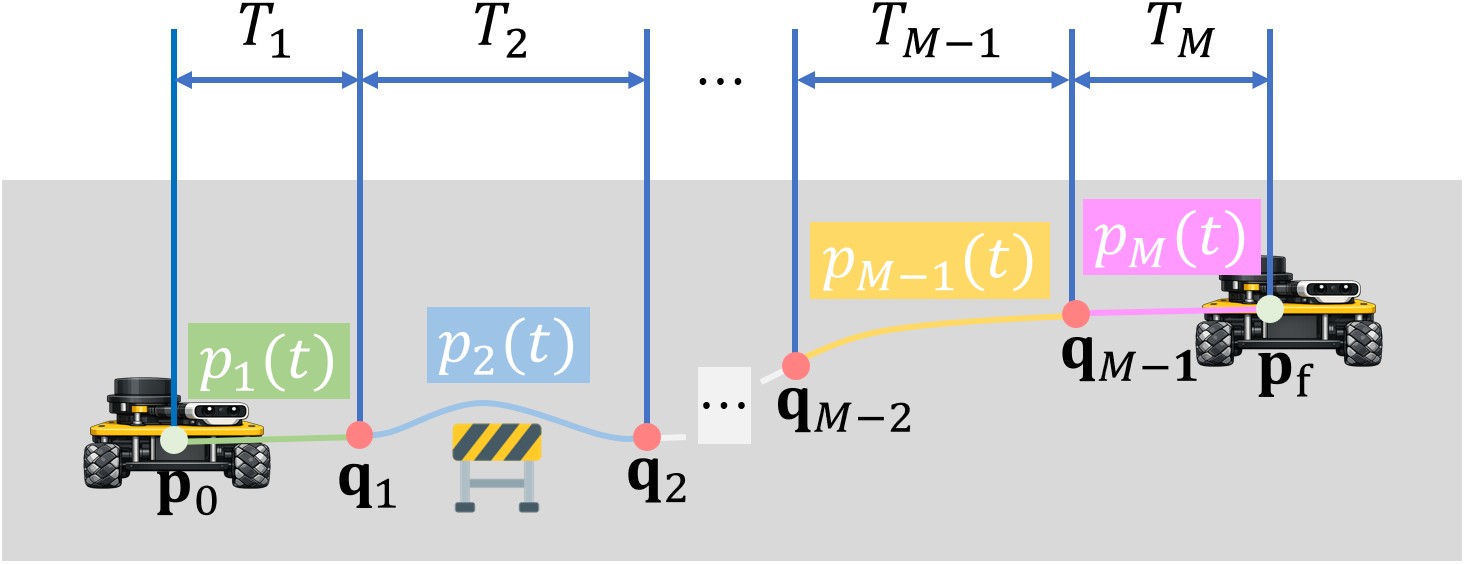}
	\vspace{-2mm}
	\caption{Schematic illustration of the $\mathfrak{T}_\mathrm{MINCO}$ trajectory representation parameterized by spatial waypoints $\mathbf{q}$ and segment durations $\mathbf{T}$.
    }
	\label{fig-MINCO}
  \vspace{-5mm}
\end{figure}

\subsection{Trajectory Representation}
\label{subsec-representation}

As illustrated in Fig.~\ref{fig-MINCO}, we adopt the minimum-control-effort trajectory representation $\mathfrak{T}_\mathrm{MINCO}$~\cite{wang2022geometrically}, parameterized by spatial waypoints $\mathbf{q}$ and segment durations $\mathbf{T}$:
\begin{gather}
\mathfrak{T}_\mathrm{MINCO}=\{\mathbf{p}(t):[0,T_{\Sigma}]\to\mathbb{R}^{m}\mid \mathbf{c}=\mathcal{M}(\mathbf{q},\mathbf{T}),\notag\\
\label{equ.minco}
\mathbf{q}\in\mathbb{R}^{m(M-1)},\quad \mathbf{T}\in\mathbb{R}_{>0}^{M}\},
\end{gather}
where $\mathbf{c}=(\mathbf{c}_1^{\top},\ldots,\mathbf{c}_M^{\top})^{\top}$ denotes the stacked polynomial coefficient matrix, $\mathbf{q}=(\mathbf{q}_1,\ldots,\mathbf{q}_{M-1})$ denotes the intermediate-point vector with $\mathbf{q}_i\in\mathbb{R}^m$, $\mathbf{T}=(T_1,\ldots,T_M)^{\top}$ denotes the segment duration vector, and $T_{\Sigma}=\sum_{i=1}^{M}T_i$ is the total trajectory duration. 

Using a third-order integrator model, we represent each trajectory by $M$ quintic ($n=5$) polynomial segments:
\begin{equation}
\mathbf{p}_i(t)=\mathbf{c}_i^{\top}\boldsymbol{\beta}(t),\quad \forall t\in[0,T_i],
\end{equation}
where $\boldsymbol{\beta}(t)=[1,t,\ldots,t^5]^{\top}$
is the polynomial basis.



\subsection {Problem Formulation}
\label{subsec.planning}


\looseness=-1
Before developing the trajectory planning method, it is necessary to first specify the state variables. 
By employing the kinematic bicycle model, WMRs are differentially flat, with the position $(x,y)$ serving as the flat outputs~\cite{tang2009differential}, from which all system states and control inputs can be uniquely determined through their finite-order derivatives.
Accordingly, the spatial dimension is given by $m=2$. By exploiting differential flatness, trajectory planning can be carried out in a low-dimensional smooth trajectory space, thereby effectively accelerating optimization~\cite{wang2022geometrically}.

\looseness=-1
We jointly optimize $\mathbf{q}$ and $\mathbf{T}$, which uniquely determine a $\mathfrak{T}_\mathrm{MINCO}$ trajectory through $\mathbf{c}=\mathcal{M}(\mathbf{q},\mathbf{T})$ for given boundary conditions. Cost gradients are computed with respect to $(\mathbf{c},\mathbf{T})$ and propagated to $(\mathbf{q},\mathbf{T})$ through the $\mathfrak{T}_\mathrm{MINCO}$ mapping in linear time~\cite{wang2022geometrically}.

\looseness=-1
Finally, we directly formulate the joint spatio-temporal trajectory planning problem with the flat outputs of WMRs as an unconstrained  optimization problem,
\begin{equation}
\label{equ-optimization}
\min_{\mathbf{q},\mathbf{T}}\;
\boldsymbol{\lambda}^{\top}
[\mathcal{P}_{\mathrm{inter}},\mathcal{P}_{\mathrm{obs}},
\mathcal{P}_{\mathrm{dyn}},\mathcal{P}_{\mathrm{form}},
\mathcal{P}_{\mathrm{ctrl}},\mathcal{P}_{\mathrm{t}},
\mathcal{P}_{\mathrm{add}}]^{\top},
\end{equation}
where the spatial variables $\mathbf{q}$ and the temporal variables $\mathbf{T}$ are jointly optimized, $\mathcal{P}_x$ represents a penalty term in the cost function, and $\boldsymbol{\lambda}$ denotes the corresponding weight vector.
Seven key factors are simultaneously optimized, including inter-robot collision avoidance ($\mathcal{P}_\mathrm{inter}$), obstacle avoidance ($\mathcal{P}_\mathrm{obs}$), dynamical feasibility ($\mathcal{P}_\mathrm{dyn}$), formation maintenance ($\mathcal{P}_\mathrm{form}$), control effort ($\mathcal{P}_\mathrm{ctrl}$), total travel time ($\mathcal{P}_\mathrm{t}$), and motion regularization $\mathcal{P}_{\mathrm{add}}$.
The corresponding penalty can be enforced by assigning a sufficiently large weight.
As an integrated formulation, JSTP~\eqref{equ-optimization} effectively balances conflicting objectives, especially formation maintenance and obstacle avoidance.
Moreover, its unconstrained form enables high-frequency trajectory planning.
Compared with optimizing spatial positions with fixed segment durations, JSTP jointly adjusts spatial positions and segment durations, thereby providing stronger spatio-temporal coordination capability and a larger solution space for handling complex and dynamic environments.

\looseness=-1
Specifically, the JSTP optimization problem (\ref{equ-optimization}) is solved using the L-BFGS algorithm~\cite{nocedal2006numerical}, a highly efficient quasi-Newton optimizer. 
In addition, continuous-time inequality constraints $g(\mathbf{p}^{(0)}(t), \ldots, \mathbf{p}^{(3)}(t)) \preceq 0$, such as dynamic feasibility constraints, are converted into finite-dimensional penalty terms by sampling a set of constraint points ${\mathbf{p}}_{i}(t_j)$ along the trajectory, where $t_j = j\cdot T_i/K_i$ is the relative time within the $i$-th polynomial piece and $K_i$ denotes the number of sampling intervals for that piece.

\subsection{Cost Functions and Gradients}
\label{subsec-cost-grad}

\looseness=-1
We then present the cost function design and gradient derivation. 
Among the considered factors, inter-robot collision avoidance ($\mathcal{P}_\mathrm{inter}$) and obstacle avoidance ($\mathcal{P}_\mathrm{obs}$) are safety-critical requirements, while dynamical feasibility ($\mathcal{P}_\mathrm{dyn}$) is a soft requirement, since final feasibility is guaranteed by the low-level controller. These three factors are essentially all range constraints, whereas the others are objectives to be minimized as much as possible.
Formation maintenance ($\mathcal{P}_\mathrm{form}$) is our primary objective, whereas control effort ($\mathcal{P}_\mathrm{ctrl}$) and total travel time ($\mathcal{P}_\mathrm{t}$) are competing objectives entailing an inherent trade-off.

\subsubsection{Obstacle Avoidance $\mathcal{P}_\mathrm{obs}$}

Obstacle avoidance is critical to safe navigation and involves both dynamic and static obstacles. 
To achieve dynamic obstacle avoidance, penalties are imposed on the trajectory points $\mathbf{p}(t_j)$ whose distance to the obstacle trajectory $\mathbf{p}_{\mathrm{obs}}(\tau_j)$ falls below the safety threshold $d_{\mathrm{thr\_obs}}$ at the corresponding global timestamp $\tau_j= \sum_{r=1}^{i-1} T_r + j\cdot T_i/K_i$. Specifically, the pointwise distance metric and obstacle penalty are defined as
\begin{gather}
d\left(\mathbf{p}(t_j), \mathbf{p}_{\mathrm{obs}}(\tau_j)\right)
=
\left\|
\mathbf{p}(t_j)-\mathbf{p}_{\mathrm{obs}}(\tau_j)
\right\|_2,
\\
g_{\mathrm{obs}}
=
\max\left\{
d_{\mathrm{thr\_obs}}^{2}
-
d^2\!\left(\mathbf{p}(t_j), \mathbf{p}_{\mathrm{obs}}(\tau_j)\right),
\,0
\right\}.
\end{gather}

The obstacle avoidance cost is then obtained by accumulating the pointwise penalties over all sampled points:
\begin{equation}
\label{equ-p-obs}
\mathcal{P}_{\mathrm{obs}}
=
\sum_{i=1}^{M}
\left(
T_i/K_i
\sum_{j=0}^{K_i}
\omega_j
\,
g_{\mathrm{obs}}^{3}
\!\left(
\mathbf{p}(t_j),
\mathbf{p}_{\mathrm{obs}}(\tau_j)
\right)
\right),
\end{equation}
where
$
(\omega_0,\omega_1,\ldots,\omega_{K_i})
=
(1/2,1,\ldots,1,1/2)
$
are the trapezoidal integration weights. 
Static obstacles can be regarded as a special case of dynamic ones, for which $\mathbf{p}_{\mathrm{obs}}(\tau_j)$ is constant, and only the nearest obstacle is considered.

For static obstacles, the gradients of $\mathcal{P}_{\mathrm{obs}}$ with respect to $\mathbf{c}_i$ and $T_i$ can be derived via the chain rule as
\begin{gather}
\frac{\partial \mathcal{P}_{\mathrm{obs}}}{\partial \mathbf{c}_i}
=
\frac{\partial \mathcal{P}_{\mathrm{obs}}}{\partial g_{\mathrm{obs}}}
\frac{\partial g_{\mathrm{obs}}}{\partial \mathbf{c}_i}
=
\frac{\partial \mathcal{P}_{\mathrm{obs}}}{\partial g_{\mathrm{obs}}}
\frac{\partial g_{\mathrm{obs}}}{\partial \mathbf{p}_i(t_j)}
\frac{\partial \mathbf{p}_i(t_j)}{\partial \mathbf{c}_i},\notag\\
\frac{\partial \mathcal{P}_{\mathrm{obs}}}{\partial T_i}
=
\frac{\mathcal{P}_{\mathrm{obs},i}}{T_i}
+
\frac{\partial \mathcal{P}_{\mathrm{obs}}}{\partial g_{\mathrm{obs}}}
\frac{\partial g_{\mathrm{obs}}}{\partial t_j}
\frac{\partial t_j}{\partial T_i},\label{equ-obs-gradients}\\
\frac{\partial \mathbf{p}_i(t_j)}{\partial \mathbf{c}_i}=\boldsymbol{\beta}(t_j), 
\frac{\partial g_{\mathrm{obs}}}{\partial t_j}
=\frac{\partial g_{\mathrm{obs}}}{\partial \mathbf{p}_i(t_j)}\,\dot{\mathbf{p}}_i(t_j), 
\frac{\partial t_j}{\partial T_i}=j/K_i,\notag
\end{gather}
where $\mathcal{P}_{\mathrm{obs},i}$ denotes the $i$-th segment's contribution to $\mathcal{P}_{\mathrm{obs}}$ in~\eqref{equ-p-obs}.
For dynamic obstacles, the gradient with respect to $T_i$ additionally needs to account for the global timestamp $\tau_j = \sum_{r=1}^{i-1} T_r + j\cdot T_i/K_i$, which is used to query the obstacle trajectories during ego-trajectory optimization.


\subsubsection{Inter-Robot Collision Avoidance $\mathcal{P}_{\mathrm{inter}}$}

Under the fully distributed planning framework, the other WMRs are essentially dynamic obstacles.
Similarly, penalties are imposed on the trajectory points $\mathbf{p}(t_j)$ whenever their distance to the trajectory points $\mathbf{p}_l(\tau_j)$ of another WMR $l\in\mathcal{R}$ is smaller than the safety threshold $d_{\mathrm{thr\_wmr}}$, where $\mathcal{R}$ denotes the set of all other WMRs.
The corresponding distance metric and pointwise penalty function are defined as follows:
\begin{gather}
\hat{d}\left(\mathbf{p}(t_j), \mathbf{p}_l(\tau_j)\right)
=
\left\|
\mathbf{E}
\left(
\mathbf{p}(t_j)-\mathbf{p}_l(\tau_j)
\right)
\right\|_2,\\
g_{\mathrm{inter}}
=
\max\left\{
d_{\mathrm{thr\_wmr}}^{2}
-
\hat{d}^{\,2}\!\left(\mathbf{p}(t_j), \mathbf{p}_l(\tau_j)\right),
\,0
\right\},
\end{gather}
where
$
\mathbf{E} := \operatorname{diag}(1,b)
$ is the weight matrix and $0<b<1$. 
Due to the nonholonomic nature of WMRs, collision avoidance should be prioritized along the lateral direction ($y$-direction). 
Therefore, $0<b<1$ is adopted to promote a larger lateral safety clearance.

$\mathcal{P}_{\mathrm{inter}}$ is constructed in the same manner as $\mathcal{P}_{\mathrm{obs}}$~\eqref{equ-p-obs}, with $g_{\mathrm{obs}}$ replaced by $g_{\mathrm{inter}}$ and an additional summation over $\mathcal{R}$.
Since $\mathcal{P}_{\mathrm{inter}}$ involves the trajectories of other WMRs, its gradient derivation needs to account for $\tau_j$, as with $\mathcal{P}_{\mathrm{obs}}$.
Owing to the unified cost function structure, the derivation is identical to that of $\mathcal{P}_{\mathrm{obs}}$.

\subsubsection{Dynamical Feasibility $\mathcal{P}_\mathrm{dyn}$}

To ensure that the planned trajectory is trackable and supports coordinated motion, dynamical constraints are imposed on the trajectory.
Violations of the velocity and acceleration limits are penalized as:
\begin{gather}
{g}_v = 
\max\{\|\dot{\mathbf{p}}(t)\|_2^2 -  v_{\mathrm{max}}^2,\, 0\},\\
{g}_a =
\max\{\|\ddot{\mathbf{p}}(t)\|_2^2 - a_{\mathrm{max}}^2,\,0\},
\end{gather}
where $v_{\mathrm{max}}$ and $a_{\mathrm{max}}$ denote the maximum speed and acceleration of the WMRs, respectively.

For the front-wheel steering constraint induced by the nonholonomic property, the steering angle $\delta$ can be obtained from the trajectory curvature $\kappa$ as $\delta= \arctan (\kappa L)$, where $\kappa = \ddot{\mathbf{p}}(t)^{\top} \mathbf{M} \dot{\mathbf{p}}(t)  / \|\dot{\mathbf{p}}(t)\|_2^3$, $L$ is the wheelbase, and $\mathbf{M}:=
\begin{bmatrix}
0 & -1\\
1 & 0
\end{bmatrix}$.
To avoid complicated trigonometric computations, the steering constraint is equivalently enforced by bounding the trajectory curvature as
$|\kappa|\leq \kappa_{\mathrm{max}}$, where
$\kappa_{\mathrm{max}}=\tan\delta_{\mathrm{max}}/L$ and $\delta_{\mathrm{max}}$ is the maximum front-wheel steering angle.
The corresponding pointwise penalty is defined as:
\begin{equation}
{g}_{\delta} = \max \left\{ \kappa^2 - \kappa_\mathrm{max}^2, 0 \right\}.
\end{equation}
The feasibility penalty $\mathcal{P}_\mathrm{dyn}$ is formulated similarly to $\mathcal{P}_\mathrm{obs}$ in~\eqref{equ-p-obs}, where $g_\mathrm{obs}$ is replaced by $({g}_v + {g}_a + {g}_{\delta})$.

\begin{table}[t]
\centering
\caption{Computation Time Comparison for Conflict-Free WMR-to-Target Assignment ($\mathrm{ms}$)}
\label{tab-runtime-comparison}
\vspace{-1mm}
\footnotesize
\setlength{\tabcolsep}{2.3pt} 
\renewcommand{\arraystretch}{1.15}
\begin{tabular}{lccccccccc}
\toprule
Formation Size & 8 & 16 & 18 & 20 & 22 & 24 & 26 & 28 & 30 \\
\midrule
Hungarian-Based~\cite{cai2023formation} & 28.4 & 26.7 & 36.5 & 37.6 & 48.2 & 43.4 & 59.4 & 55.5 & 75.0 \\
TCF-R2T (Ours) & \textbf{0.7} & \textbf{4.4} & \textbf{5.0} & \textbf{8.7} & \textbf{9.1} & \textbf{14.8} & \textbf{15.4} & \textbf{22.8} & \textbf{24.3} \\
Reduction ($\%$)  & 97.5 & 83.5 & 86.3 & 76.9 & 81.1 & 65.9 & 74.1 & 58.9 & 67.6 \\
\bottomrule
\end{tabular}
\end{table}

\subsubsection {Formation Maintenance $\mathcal{P}_\mathrm{form}$}

As described in Section~\ref{subsec-description}, the WMR formation is characterized by the Laplacian matrix $\tilde{\mathbf{L}}$. Accordingly, we define the formation error metric $f_{\mathrm{e}}$ as follows, based on the difference between the current and desired formation Laplacians:
\begin{equation}
\label{equ.formation_function}
f_\mathrm{e}= \| \tilde{\mathbf{L}} - \tilde{\mathbf{L}}^* \|_F^2 = \mathrm{tr} \{ (\tilde{\mathbf{L}} - \tilde{\mathbf{L}}^*)^{\top} (\tilde{\mathbf{L}} - \tilde{\mathbf{L}}^*) \},
\end{equation}
where $\mathrm{tr}\{ \cdot \}$ denotes the trace operator, $\| \cdot \|_F$ denotes the Frobenius norm, and $\tilde{\mathbf{L}}$ and $\tilde{\mathbf{L}}^*$ are the Laplacian matrices characterizing the current and desired formations, respectively. 
Since $f_{\mathrm e}$ is differentiable, it can be directly incorporated into the optimization.
The formation penalty is then defined as $g_\mathrm{form}=f_\mathrm{e}\!\left(\mathbf{p}(t_j), \{\mathbf{p}_l(\tau_j)\}_{l\in\mathcal{R}}\right)$, where $\mathcal{R}$ denotes the set of all other WMRs, and $\tilde{\mathbf{L}}$ is computed from these positions as described in Section~\ref{subsec-description}. $\mathcal{P}_{\mathrm{form}}$ is then obtained by replacing $g_{\mathrm{inter}}$ in $\mathcal{P}_{\mathrm{inter}}$ with $g_{\mathrm{form}}$.
The gradients of $\mathcal{P}_{\mathrm{form}}$ with respect to $\mathbf{c}_i$ and $T_i$ follow the same derivation as those of $\mathcal{P}_{\mathrm{inter}}$.


\subsubsection{Control Effort $\mathcal{P}_\mathrm{ctrl}$}

The control effort is directly measured by the integral of squared jerk,
which is differentiable and inherently serves as a trajectory smoothness cost.

\subsubsection{Total Time $\mathcal{P}_\mathrm{t}$}
The travel time cost is directly given by \( \mathcal{P}_\mathrm{t} = \sum_{i=1}^M T_i \). Since time is an independent parameter in $\mathfrak{T}_\mathrm{MINCO}$, the gradients can be directly derived as 
${\partial \mathcal{P}_\mathrm{t}}/{\partial \mathbf{c}_i} = \mathbf{0}, \, {\partial \mathcal{P}_\mathrm{t}}/{\partial {T}_i} = 1$. 
In implementation, we use a positive-valued mapping to ensure $T_i>0$.
\subsubsection{Motion Regularization $\mathcal{P}_{\mathrm{add}}$}

The graph-based formation cost does not uniquely determine formation orientation.
We therefore penalize centroid-relative position errors:
\begin{equation}
\label{equ-relative-position}
g_{\mathrm{pos}}=\frac{1}{N}\sum_{u\in\mathcal{V}}
\left\|(\mathbf{p}_u-\bar{\mathbf{p}})-(\mathbf{r}_u-\bar{\mathbf{r}})\right\|_2^2,
\end{equation}
where $\mathbf{p}_u$ denotes the position of WMR
$u\in\mathcal{V}$, $\mathbf{r}_u$ is its TVFR reference from~\eqref{eq:smooth_formation_reference}, and overbars denote team averages. All quantities are evaluated at $\tau_j$, using $\mathbf{p}_i(t_j)$ for the ego WMR and received trajectories for the other WMRs.
A relative-velocity cost tracks the TVFR derivatives to reduce repeated acceleration and braking.
Since relative costs cannot prevent collective drift, lateral route anchoring penalizes centroid position error and velocity in the lateral direction during transitions along straight route segments.
The weighted terms are integrated into $\mathcal{P}_{\mathrm{add}}$; details are available on the \href{https://dongjh20.github.io/CFOO}{\textcolor{blue}{project website}}.
\begin{figure}[t]
	\centering
    \setlength{\subfigtopskip}{0pt}
    \setlength{\subfigbottomskip}{0pt}
    \setlength{\subfigcapskip}{-3pt}
	\subfigure[Experimental scenario]
	{\includegraphics[width=0.36\textwidth]{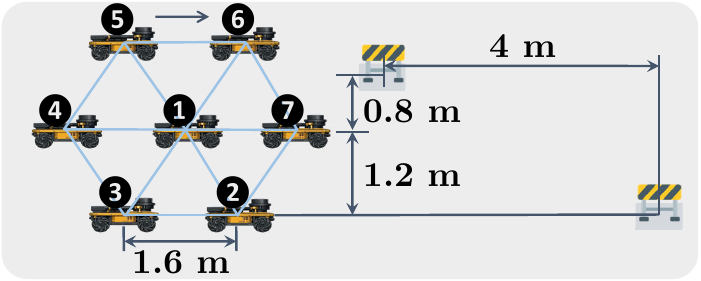}
		\label{fig-maintenance-scenario} }
  \par\vspace{3pt}
	\subfigure[Experimental result]
	{\includegraphics[width=0.4\textwidth]{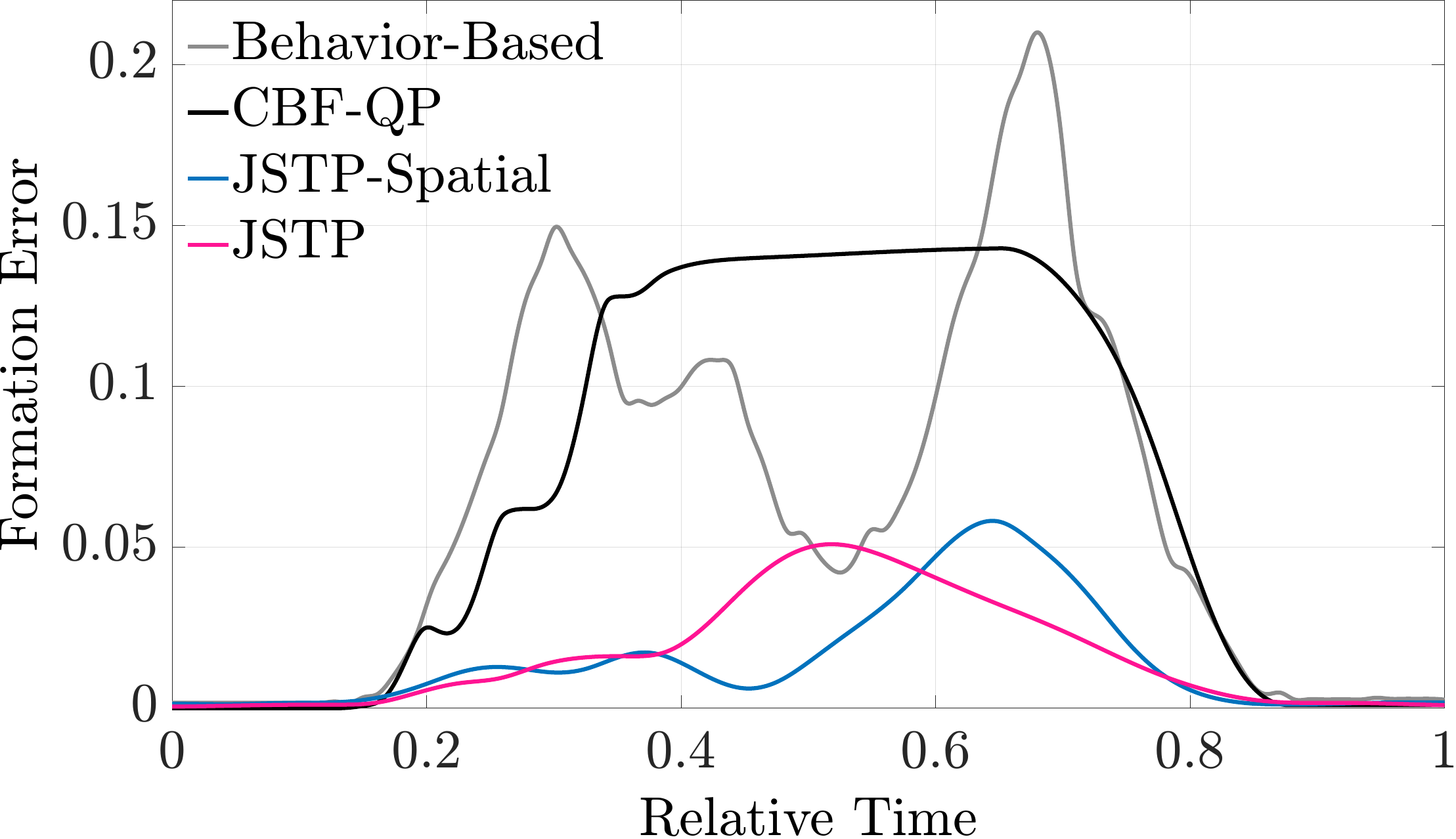}
		\label{fig-maintenance-error}}
	\par\vspace{2pt}
	\caption{Comparison of formation maintenance  performance. (a) Experimental scenario, where seven WMRs are uniformly arranged in an interlaced three-column formation, and two consecutive obstacles are placed ahead. (b) Experimental results, which demonstrate that our method achieves the lowest peak formation error and smaller error fluctuations under consecutive obstacle disturbances.}
	 \vspace{-4mm}
	\label{fig-maintenance }
\end{figure}

\begin{figure*}[t]
	\centering
    \setlength{\subfigcapskip}{-3pt}
\setlength{\subfigtopskip}{0pt}
\setlength{\subfigbottomskip}{-2pt}

\subfigure[\strut Experimental trajectories]{%
    \includegraphics[width=0.83\textwidth]{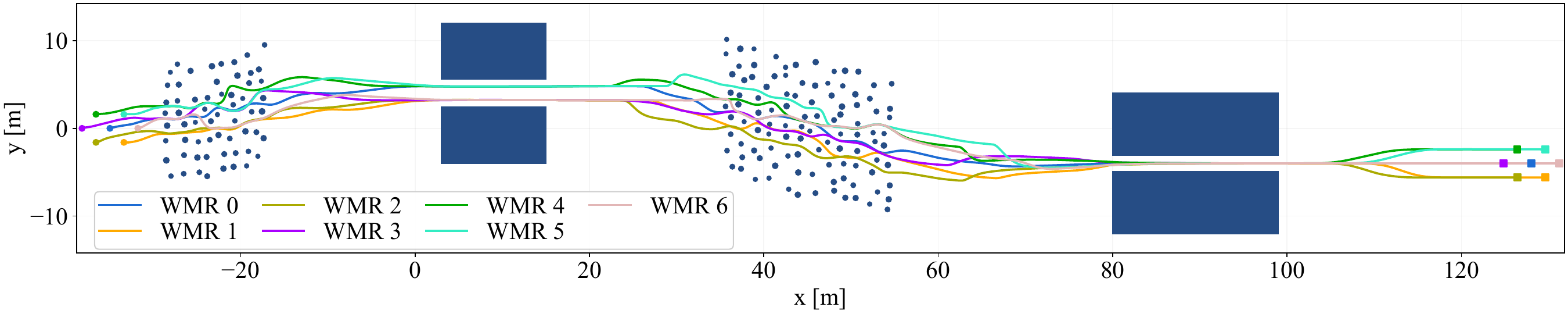}%
    \label{fig-full_trajectory}%
}
\par\vspace{4.2pt}

\subfigure[\strut Formation error]{%
    \includegraphics[width=0.83\textwidth]{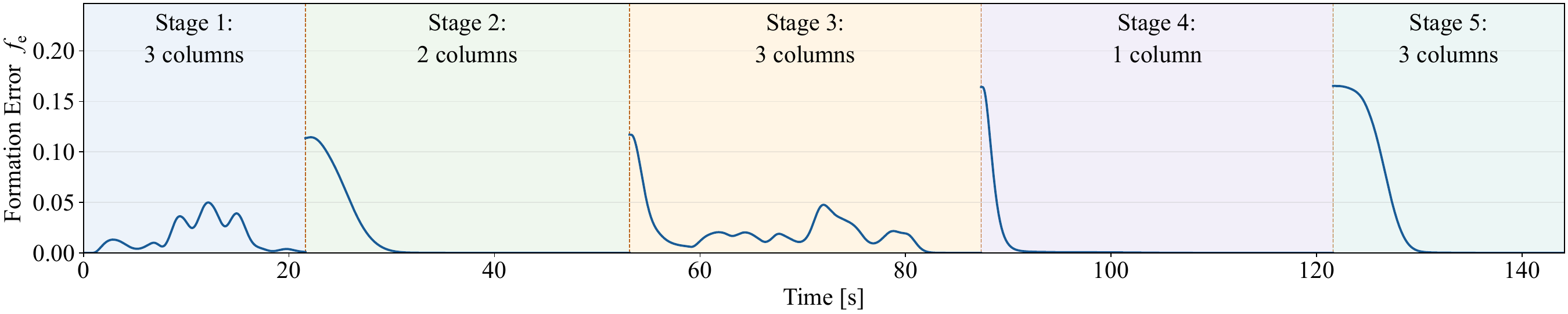}%
    \label{fig-formation_error_all}%
}
\par\vspace{4.2pt}

\subfigure[\strut Transition trajectory comparison: 3→2 columns]{%
    \includegraphics[width=0.41\textwidth]{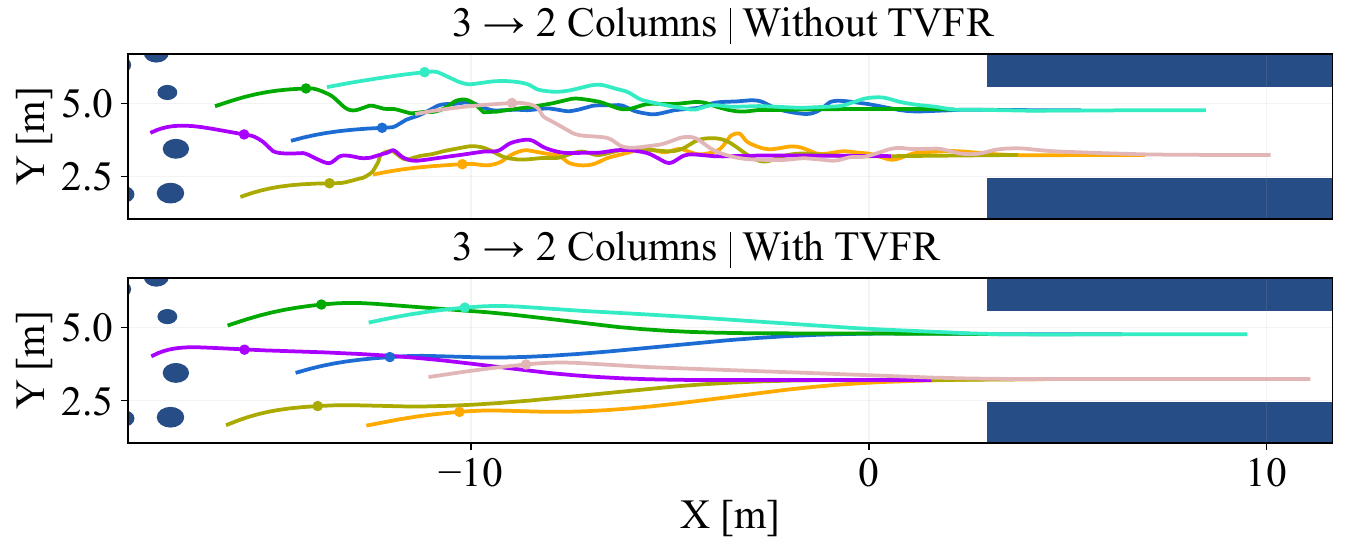}%
    \label{fig-tvfr_32_trajectory}%
}
\subfigure[\strut Lateral velocity comparison: 3→2 columns]{%
    \includegraphics[width=0.41\textwidth]{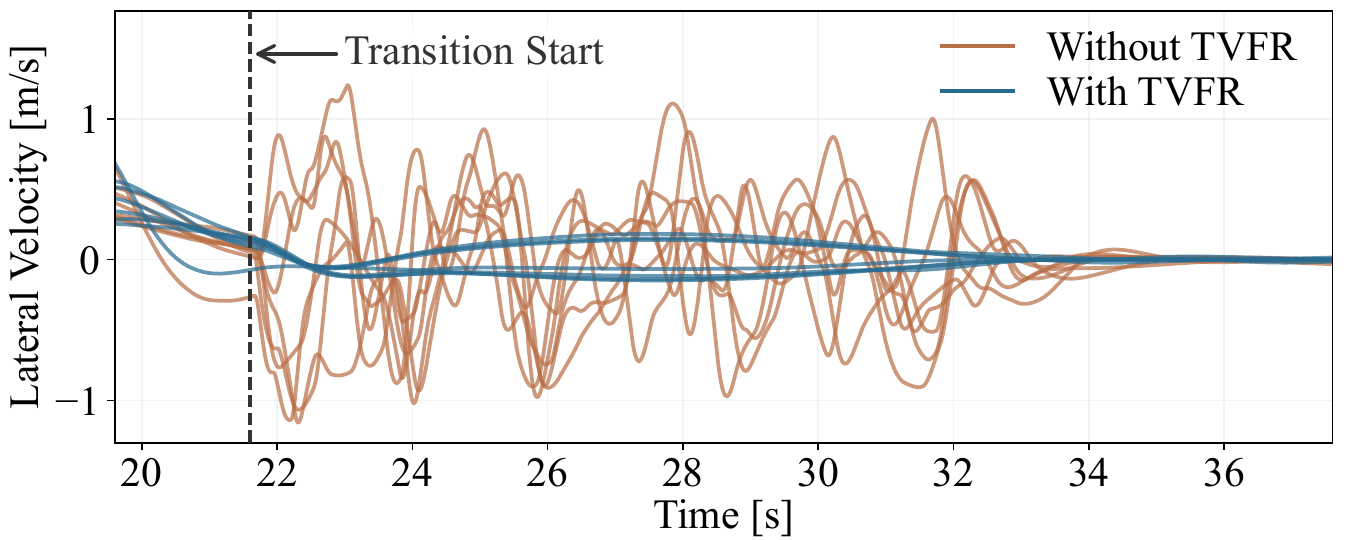}%
    \label{fig-tvfr_32_lateral_velocity}%
}
\par
	\caption{Continuous formation navigation in simulation; see Fig.~\ref{fig-maintenance-scenario} for WMR IDs. (a) WMR trajectories through dense obstacles and narrow passages, with formation transitions following a three--two--three--one--three-column sequence. (b) Formation error $f_\mathrm{e}$~\eqref{equ.formation_function} across the five stages. (c) Trajectories during the three-to-two-column transition without TVFR (top) and with TVFR (bottom). (d) Lateral velocity comparison during the three-to-two-column transition.
    }
	 \vspace{-2mm}
	\label{fig-full_sim }
\end{figure*}

\begin{figure*}[t]
	\centering
	\includegraphics[width=0.83\textwidth]  {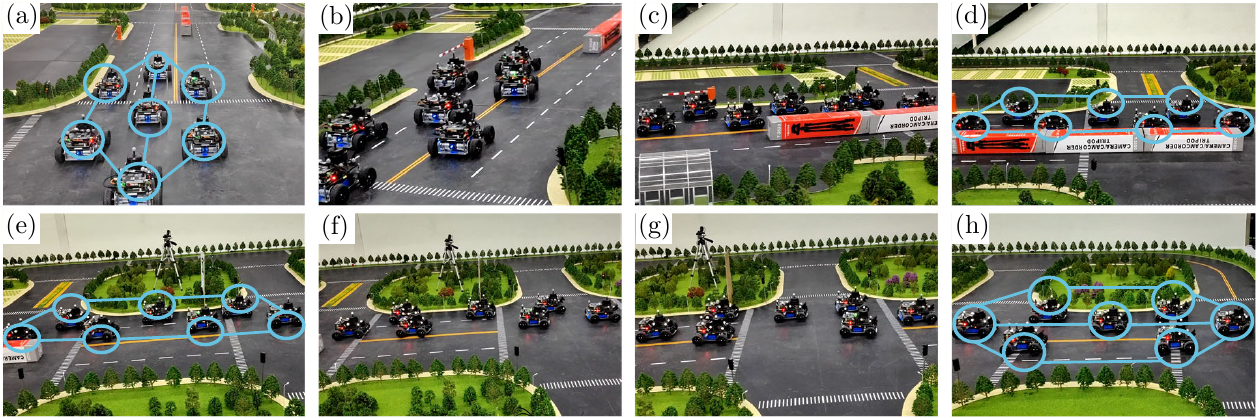 }
	\vspace{-3mm}
	\caption{Snapshots of our real-world continuous formation navigation experiment. Due to changes in the navigable area, the WMR formation transitions from three columns to two columns and back to three columns. The experiment demonstrates TCF-R2T’s ability to promptly generate conflict-free assignments and the effectiveness of CFOO in enabling continuous navigation in varying environments. Experimental videos are available on our \href{https://dongjh20.github.io/CFOO}{\textcolor{blue}{project website}}.
    }
	\label{fig-exp-transition}
  \vspace{-4mm}
\end{figure*}

\section{Experiments}
\label{sec-experiment}

\looseness=-1
In this section, we first present comparative results on formation generation efficiency and formation maintenance performance.
We then demonstrate the effectiveness and adaptability of CFOO through both simulation and real-world experiments. 
Experimental videos and comparative results with and without the TVFR mechanism are available on the 
\href{https://dongjh20.github.io/CFOO}{\textcolor{blue}{project website}}.

\subsection{Formation Generation Efficiency}

\looseness=-1
Since formation geometry is generated directly from the navigable width, we evaluate formation generation efficiency through the computation time for conflict-free assignment.
Here, we compare the runtime of TCF-R2T and the classical Hungarian-based method~\cite{cai2023formation} for computing conflict-free assignments across different formation sizes, ranging from $8$ to $30$ WMRs.
The scenario involves a formation transition from four columns to two columns caused by changes in the navigable area.
The results are summarized in Table~\ref{tab-runtime-comparison}, which shows that TCF-R2T reduces the runtime by at least $55\%$ across all formation sizes.
This improvement is attributed to the constructed time-expanded network, which resolves motion conflicts at each time step and thus enables the optimal assignment to be computed directly.
In contrast, Hungarian-based methods~\cite{cai2023formation} require iterative processes of assignment attempts and conflict resolution.


\subsection{Formation Maintenance Performance}

\looseness=-1
We compare our JSTP with methods that can simultaneously handle formation maintenance and obstacle avoidance, including the behavior-based method with unit-center reference~\cite{balch1998behavior}, the control barrier function–based quadratic programming method (CBF-QP)~\cite{wang2017safety}, and JSTP-Spatial, a variant of JSTP that optimizes spatial variables while keeping segment durations fixed.
Considering the scenario shown in Fig.~\ref{fig-maintenance-scenario}, a seven-WMR formation arranged in three columns encounters two consecutive obstacles: one located between two columns and the other directly in front of the third column. 
This scenario captures a representative obstacle pattern in obstacle-rich environments.
The formation error metric $f_\mathrm{e}$ is adopted for comparison, and the symmetrically normalized Laplacian matrix is used to eliminate the influence of spatial scale.
The results are presented in Fig.~\ref{fig-maintenance-error}, which demonstrate that our method achieves the lowest peak formation error and smaller error fluctuations under consecutive obstacle disturbances. Specifically, the peak formation error is reduced by $75\%$ compared with the behavior-based method and by $64\%$ compared with CBF-QP. Moreover, compared with JSTP-Spatial, JSTP further reduces the peak formation error by $13\%$, demonstrating the benefit of jointly optimizing spatial positions and segment durations.

\subsection{Simulation and Real-World Experiments}

\looseness=-1
The complete experimental videos are available on our \href{https://dongjh20.github.io/CFOO}{\textcolor{blue}{project website}}, and representative results are presented below.
Fig.~\ref{fig-full_trajectory} presents a five-stage simulation of continuous formation navigation. 
Each WMR independently perceives obstacles using simulated LiDAR and sends the perceived information to the formation manager.
The WMRs navigate through dense obstacles in three columns, transition to two columns to traverse a narrow passage, return to three columns through another dense obstacle field, transition to a single column through an exceptionally narrow passage, and finally resume a three-column formation. 
The corresponding formation errors are shown in Fig.~\ref{fig-formation_error_all}.
These results demonstrate CFOO's effectiveness in enabling continuous formation navigation in varying environments: TCF-R2T promptly generates conflict-free assignments, while JSTP coordinates WMR motion for efficient, collision-free formation transitions and navigation through dense obstacles. 
We also conduct an ablation study to assess the effectiveness of the TVFR mechanism under the same setup as the five-stage simulation. The results are shown in Figs.~\ref{fig-tvfr_32_trajectory} and~\ref{fig-tvfr_32_lateral_velocity}, which demonstrate that TVFR can effectively suppress motion oscillations caused by abrupt updates to JSTP’s desired formation.

\looseness=-1
In the real-world experiments, seven Ackermann-steered WMRs ($0.22\,\mathrm{m} \times 0.19\,\mathrm{m} \times 0.13\,\mathrm{m}$) are deployed in a $9\,\mathrm{m} \times 5\,\mathrm{m}$ area. Markers attached to the WMRs and obstacles enable pose tracking via the FZMotion motion capture system, while real-time speeds are measured by onboard sensors. 
Each WMR 
communicates with the formation manager running on an external host via Robot Operating System (ROS) messages.
Fig.~\ref{fig-exp-transition} shows snapshots from the experiment. As the navigable area narrows, the WMRs promptly transition from three columns to two columns without conflicts and smoothly return to three columns after traversing the narrow passage. 
The real-world experiments demonstrate CFOO's ability to adapt WMR formations to varying environments while maintaining smooth, continuous navigation.

\raggedbottom
\section{Conclusion}
\label{sec-conclusion}

\looseness=-1
In this paper, we propose CFOO, an online formation orchestration framework for continuous formation navigation of WMRs in varying environments. 
It consists of centralized conflict-free formation generation and distributed robust formation maintenance. 
Specifically, we develop the TCF-R2T algorithm for rapid, conflict-free WMR-to-target assignment by resolving motion conflicts at each time step in the time-expanded network, enabling timely generation of environment-adaptive formations.
Meanwhile, we develop JSTP to jointly optimize spatial positions and temporal durations, thereby enhancing coordination among WMRs and improving formation maintenance performance under obstacle disturbances.
Moreover, we design the TVFR mechanism for smooth transitions between successive desired formations.
Comparative results demonstrate the advantages of our proposed methods, while simulation and real-world experiments validate the effectiveness and adaptability of CFOO. 

\looseness=-1
In the future, we plan to integrate the generalizable understanding and reasoning capabilities of large language models (LLMs) into CFOO to extend the architecture to accommodate changing task requirements, particularly during formation generation.
We also intend to deploy and evaluate CFOO in a broader range of environments, such as scenarios with structured road constraints and complex mountainous terrains with significant elevation changes.
Finally, we are also interested in extending CFOO to other robotic platforms, such as unmanned aerial vehicles (UAVs).

\bibliographystyle{IEEEtran}
\begingroup
\let\originalbibitem\bibitem
\renewcommand{\bibitem}[1]{\originalbibitem{#1}\looseness=-1\interlinepenalty=10000}
\bibliography{IEEEabrv,mybibfile}
\endgroup

\end{document}